\documentclass[conference]{IEEEtran}
\IEEEoverridecommandlockouts
\usepackage{cite}
\usepackage{amsmath,amssymb,amsfonts}
\usepackage{algorithmic}
\usepackage{graphicx}
\usepackage{textcomp}
\usepackage{xcolor}

\usepackage{multirow}
\usepackage{multicol}
\usepackage{caption}
\usepackage{booktabs}
\usepackage{makecell}
\usepackage{marvosym}

\def\BibTeX{{\rm B\kern-.05em{\sc i\kern-.025em b}\kern-.08em
    T\kern-.1667em\lower.7ex\hbox{E}\kern-.125emX}}

\begin{document}

\title{Eliminating the Language Bias for Visual Question Answering with fine-grained Causal Intervention\\
}

\author{\IEEEauthorblockN{Ying Liu*, Ge Bai*, Chenji Lu*, Shilong Li, Zhang Zhang, Ruifang Liu and Wenbin Guo\textsuperscript{\Letter}\thanks{{\Letter} Corresponding author}}
\IEEEauthorblockA{\textit{Beijing University of Posts and Telecommunications} \\
Beijing, China \\
\{ly1422615223, white, luchenji1, lishilong2019210645, zzpal, lrf, gwb\}@bupt.edu.cn}
\thanks{$^*$The first three authors contribute equally.}
% \and
% \IEEEauthorblockN{2\textsuperscript{nd} Ge Bai}
% Beijing, China \\
% white@bupt.edu.cn}
% \and
% \IEEEauthorblockN{3\textsuperscript{rd} Lu chenji}
% Beijing, China \\
% luchenji1@bupt.edu.cn}
% \and
% \IEEEauthorblockN{4\textsuperscript{th} Shilong Li }
% Beijing, China \\
% lishilong2019210645@bupt.edu.cn}
% \and
% \IEEEauthorblockN{5\textsuperscript{th} Zhang Zhang}
% Beijing, China \\
% zzpal@bupt.edu.cn}
% \and
% \IEEEauthorblockN{6\textsuperscript{th} Ruifang Liu}
% Beijing, China \\
% lrf@bupt.edu.cn}
% \and
% \IEEEauthorblockN{6\textsuperscript{th} Wenbin Guo }
% Beijing, China \\
% gwb@bupt.edu.cn}
}

\maketitle

\begin{abstract}
Despite the remarkable advancements in Visual Question Answering (VQA), the challenge of mitigating the language bias introduced by textual information remains unresolved. Previous approaches capture language bias from a coarse-grained perspective. However, the finer-grained information within a sentence, such as context and keywords, can result in different biases. Due to the ignorance of fine-grained information, most existing methods fail to sufficiently capture language bias. 
In this paper, we propose a novel causal intervention training scheme named CIBi to eliminate language bias from a finer-grained perspective. Specifically, we divide the language bias into context bias and keyword bias. We employ causal intervention and contrastive learning to eliminate context bias and improve the multi-modal representation. Additionally, we design a new question-only branch based on counterfactual generation to distill and eliminate keyword bias. Experimental results illustrate that CIBi is applicable to various VQA models, yielding competitive performance.
\end{abstract}


\section{Introduction}
\label{sec:intro}
Visual Question Answering (VQA)\cite{SAN,UpDn} is a challenging multi-modal task. 
Given a question and an image, the model's extractor\cite{VGG,LSTM} extracts the textual and visual features independently, which are then used for joint reasoning.
However, recent studies\cite{VQACP,RUBi} have shown that most existing VQA models over-rely on superficial correlations between certain types of questions and answers, rather than truly comprehending both textual and visual cues.
One straightforward solution\cite{2022rethinking,SCR} to mitigate language bias is to enhance the training data by using extra annotations or data augmentation, which requires high costs. 
Some existing methods\cite{RUBi,question-only,LPF,AdaVQA} attempt to mitigate language bias by designing complex model branches, which are utilized during the training phase to capture biases introduced by the textual modality. 
Besides, counterfactual generation based methods\cite{CF-VQA,CSS,2022efficient,Mutant} result in significant performance improvements compared to other debiasing methods by balancing the training data. 
These methods generate counterfactual inputs by fully masking the question or partially masking keywords from a coarse-grained perspective. 
However, due to the existence of fine-grained language bias in syntactic structure, keyword and context, most existing methods either fail to sufficiently capture language bias or overly eliminate the textual information, leading to the loss of some useful textual features. 
For instance, in VQA v1.0\cite{VQAv1}, approximately 80\% of questions containing “banana” have the answer “yellow”, which we refer to as keyword bias. And about 90\% of questions starting with “Do you see a” have the answer “yes”\cite{CF-VQA}, which we refer to as syntactic structure bias. Additionally, almost 80\% specific pattern of question-type words and keyword such as “What color” + “plate” have the answer “white”, which we refer to as context bias.

Inspired by this, we propose a fine-grained training scheme named CIBi to eliminate language bias from a causal perspective\cite{pearl2016book,SCM}. 
Different from the previous coarse-grained debiasing methods, we define the language bias as the confounder bias and further divide it into syntactic structure, keyword and context bias. 
Then we eliminate fine-grained biases that exist in syntactic structure, keyword and context with causal intervention at token-level. 
For context bias, 
we generate two corresponding counterfactual samples for each original VQA training sample based on two synthesizing mechanisms. Specifically, we synthesize counterfactual questions by replacing keywords and syntactic structure with semantically similar synonyms, respectively. The original image and counterfactual question compose a new VQ pair. Then we utilize contrastive learning to induce the model to learn a unbiased multi-modal representation using counterfactual samples.
For syntactic structure and keyword bias, we generate counterfactual questions by masking certain keywords and syntactic structure. By training the keywords and syntactic structure debiasing branch, we distill the effect of keywords and syntactic structure bias at token-level and subtract it from the total causal effect. 
Overall, we have eliminated language bias from a fine-grained perspective. It is worth noting that each branch can be considered as a plug-and-play auxiliary module directly integrated into existing VQA models to enhance their unbiased reasoning capabilities. 

We summarize our main contributions as follows:

\begin{itemize}

    \item We explain the causes of bias at a fine-grained level and formulate the VQA debiasing process from a causal perspective.
    
    \item We propose a generic training scheme called CIBi to eliminate syntactic structure, keyword and context bias respectively with causal intervention.
    
    \item Experimental results demonstrate the effectiveness of our CIBi training scheme, as it proves adaptive across various VQA models, leading to enhanced model performance.
\end{itemize}

\section{Related Work}
\subsection{Language Bias in VQA}
%\textbf{1) Language Bias in VQA.} 
The Visual Question Answering (VQA) task takes images and textual questions as input, aiming to make prediction based on both textual and visual features. Current research\cite{RUBi,VQACP,LPF} shows that VQA models often tend to overly rely on language biases present in the dataset for answer reasoning, resulting in lower generalization performance. 
When the most of the answers to the question “What color are the bananas” are ”yellow”, VQA models learn from the statistical regularities between the most occurring answer yellow and certain patterns “what color” in the question, neglecting visual modality from the images\cite{RUBi}. To better address language bias problem in VQA, Agrawa propose the VQA-CP\cite{VQACP} dataset in which training and test sets have different distributions. The performance of many conventional VQA models drop significantly on VQA-CP due to distributional differences.

\subsection{Debiasing Strategies in VQA}
% \textbf{2$)$Debiasing Strategies in VQA.} 
Most of recent solutions to reduce the language bias in VQA can be grouped into two categories. One straightforward solution is to create more balanced training data by implicit/explicit data argumentation. For example, Zhang $et.al.$\cite{Yinandyang} collected complementary abstract scenes with opposite answers for all binary questions. And Goyal $et.al.$\cite{MakingVMatter} extended this idea into real images and all types of questions. Das\cite{HumanAttention} and Park\cite{Multimodalexplanations} exploited human visual and textual explanations respectively to strengthen the visual grounding in VQA.
Another solution is to design a separated QA branch to capture and eliminate the language prior which can further be grouped into two types: adversary-based and ensemble-based. Adversary-based methods\cite{AdversarialRegularization} train the question-only branch to weakened unwanted correlations between questions and answers in an adversarial training scheme. Ensemble-based methods\cite{RUBi,LPF} employ an ensemble strategy to combine the predictions of two models, deriving training gradients based on the fused answer distributions.

\subsection{Causal Inference}
%\textbf{3$)$Causal Inference.} 
Recently, causal inference\cite{pearl2016book} has been applied to many tasks of natural language processing and computer vision, and it shows promising results and provides strong interpretability and generalizability, including text generation, language understanding, visual explanations, image recognition, zero-shot and few-shot learning, representation learning, semantic segmentation, and visual Question
Answering tasks. In computer vision tasks, previous efforts have tackled the problem of imbalanced data distribution through counterfactual intervention, causal effect disentanglement, the back-door and front-door adjustments. Different from the previous methods, our method utilizes fine-grained causal inference to generate counterfactual input.
%In natural language processing tasks, recent causal models have been applied to various tasks such as text generation and language understanding generated counterfactuals for weakly supervised tasks, namely entity recognition (NER) and text classification, by replacing the target entity with another entity or its antonyms, respectively. Nan et al. (2021) mitigated spurious correlations in long-tailed label distribution for information extraction tasks through counterfactual generation.

\section{METHODOLOGY}
\label{method}
\begin{figure*}[!t]
\centering
\resizebox{1.0\textwidth}{!}{\includegraphics{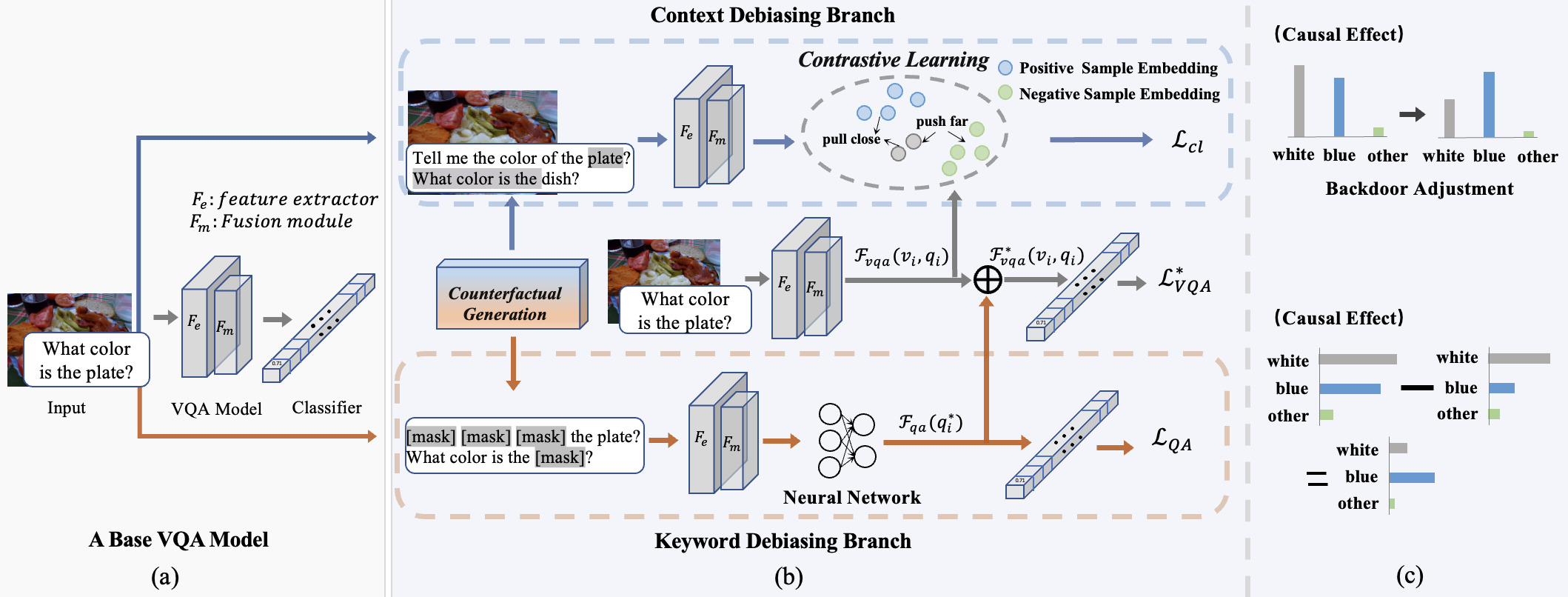}}
\caption{An overview of CIBi. (a) shows the architecture of a base VQA model. (b) illustrates the training scheme of CIBi. (c) shows our cause-effect look at language bias in VQA.}
\label{fig:model}
\end{figure*}

%Given input image �� , visual emotion recognition aims to solve the problem �� (�� |�� ), where �� is the emotion label. In this section, we will detail how confounder �� undermines the objective of �� (�� |�� ) (Section 3.1) to raise dataset bias, how such a bias effect can be removed (Section 3.2), and what our unbiased solution is (Section 3.3).
Fig.\ref{fig:model} gives an overview of CIBi, which is composed of four parts: base VQA model, context debiasing branch, syntactic structure and keyword debiasing branch and classifier. In this section, we introduce the structural causal model for VQA (Section \ref{Sec:Structural Causal Model}) and debiasing training scheme CIBi (Section 3.3, 3.4, 3.5, 3.6)
%Given a multi-modal dataset $D = \{V_i, Q_i, A_i\}^N_i$  including a picture $V_i \in V$, a question $Q_i \in Q$, and an answer $A_i \in A$, the task of VQA model is to learn a mapping function $F_{vqa}: V \times Q \rightarrow R^c$. 

\label{Sec:SCM}
\begin{figure}[!t]
    \centering
    \includegraphics[width=0.95\columnwidth]{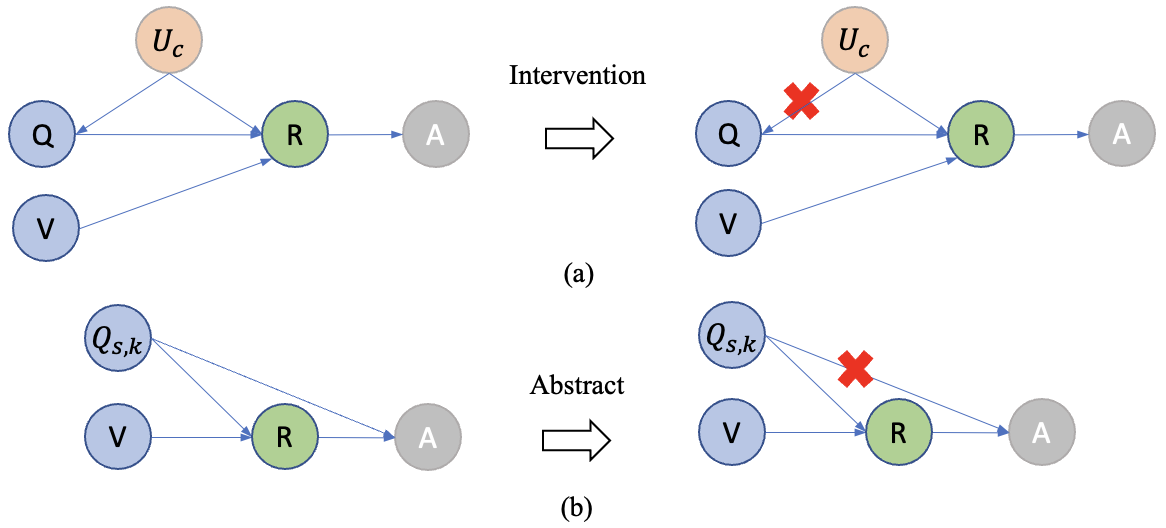}
%因果图，图(2)描述了VQA的结构因果模型，其中图(a)表示理想的表征学习过程，图(b)表示实际的带混淆偏差的表征学习过程，图(c)中展示了混淆因素、干预操作以及我们尝试进行去混淆的方法的因果图。
    \caption{Causal graphs of VQA model.}
    %\caption{Causal graphs of VQA model. (a) portrays the ideal representation learning process, (b) demonstrates the practical representation learning process with confounding bias, and (c) illustrates confounder, $do$-operator and sketch causal graphs of our attempts to de-confounder.}
    \label{SCM}
\end{figure}

%模型图

\subsection{Task Definition}
\label{Sec:CIBi}
%为了克服VQA中的语言偏见，我们提出了基于元素干预的去混淆偏差网络（CIBi）来处理VQA任务，通过嵌入因果调整和路径阻断的方法。图3概述了CIBi，由四个部分组成：骨干网络、因果分离、混淆因素块和分类器。backbone 指的是一般的vqa model, 因果分离部分进行因果干预将Qt和Uwc之间的因果效应分开，增强useful question feature的因果效应，混淆蒸馏模块进行反事实干预，分类器预测答案。
%Compared with heuristic debiasing methods like building larger datasets, CIBi is designed based on well-studied deconfounding theory. 
In accordance with the common formulation\cite{RUBi}, we define the VQA task as a multi-class classification problem. Given a multi-modal dataset $D = \{V_i,Q_i,A_i\}^N_i$ where each sample is a triplet, including a picture $V_i \in \mathcal {V}$, a question $Q_i \in \mathcal {Q}$, and an answer $A_i \in A$, the task of VQA model is to learn a mapping function $\mathcal{F}_{vqa}:\mathcal {V} \times \mathcal {Q} \rightarrow \mathbb{R}^{|A|}$. For each image $V_i$, visual features $v_i$ are extracted by an image encoder $E_v:\mathcal {V} \rightarrow \mathbb{R}^{n_v\times d_v}$ to generate a set of $n_v$ vectors of dimension $d_v$. 
For each question $Q_i$, textual features $q_i$ are extracted by a question encoder $E_q:\mathcal {Q} \rightarrow \mathbb{R}^{n_q\times d_q}$ to generate a set of $n_q$ vectors of dimension $d_q$. 
The final answer distribution is predicted by a mapping function $\mathcal{F}_{vqa}(v_i,q_i)$. The classical learning strategy of VQA models is to minimize the standard cross-entropy criterion over a dataset of size $N$:

\begin{equation}
    \setlength{\abovedisplayskip}{3pt}
    \setlength{\belowdisplayskip}{3pt}
    %\small
    \mathcal{L}_{VQA} = -\frac{1}{N}\sum^{N-1}_{i=0}log(softmax(\mathcal{F}_{vqa}(v_i,q_i))) [A_i]
\end{equation}
%The Backbone refers to a generic VQA model framework, such as S-MRL\cite{RUBi}, UpDn\cite{UpDn}. The Causal-effect Adjustment module separates the causal effects between $Q_t$ and $U_{w,c}$, enhancing the causal effects of useful question features. The Confounder Distillation module conducts counterfactual generations for confounding removal. The Classifier is responsible for answer prediction. Additionally, $F_e$,$M_f$ refer to feature extractor and Multi-modal fusion, respectively.

%骨干网络由特征提取和多模态特征融合两个模块组成，其中特征提取包括现有的图像特征提取器（例如，VGG [引用] 或 ResNet [引用]）以及文本特征提取器（例如，[引用]）。
%The backbone network consists of two parts: feature extraction and multi-modal feature fusion. The feature extraction includes existing image feature extractors and text feature extractors. We use image feature extractors 

\subsection{Structural Causal Model for VQA}
\label{Sec:Structural Causal Model}
%这部分参考的文献：Counterfactual VQA: A Cause-Effect Look at Language Bias；Two Causal Principles for Improving Visual Dialog
%结构因果模型（SCM）旨在分析因果关系。therefore，我们运用因果理论，构建因果图，分析变量之间的因果效应，指导我们设计模型。正如我们在第1节中所讨论的，VQA训练数据中存在keyword bias and context bias，这些特殊的关键词（banana）和语境（do you see a）存在于VQA数据生成过程中，是数据分布不均导致的，它同时影响着模型的问题Q和答案A。We consider Uw,c as an unobserved confounder for A. 输入（Q， V）与答案之间的相关性部分或完全来于confounder固定的语境句式或者关键词
%Q表示文本模态信息，Q*表示被mask了关键词的模态信息，V表示视觉模态信息，R表示多模态表征，A表示需要预测的答案。
%context bias存在在句法结构中，通过影响多模态表征R的学习来对影响答案A的产生，keyword bias直接对A产生影响，所以我们对这两种bias的处理和建模方式不同。
Structural Causal Models (SCM) are designed for causal analysis. As shown in Figure \ref{SCM}, $Q$, $V$, $R$, and $A$ represent textual modality information, visual modality information, multimodal representation and answer, respectively. Context bias exists in syntactic structures and keywords, influencing the learning of the multi-modal representation $R$ and subsequently affecting the generation of the answer $A$. We consider high-frequency context patterns as confounder $U_c$ and context bias as confounder bias. 
The existence of confounder $U_c$ enables a backdoor path\cite{pearl2016book} between $Q$ and $R$, making them spuriously correlated even if there is no direct causality between them. As shown in Figure \ref{SCM}a, where the answer $A$ is influenced by three paths: $V \rightarrow R \rightarrow A$, $Q \rightarrow R \rightarrow A$, and $U_{c} \rightarrow R \rightarrow A$.
To mitigate context bias, we block the backdoor path by applying the backdoor adjustment theorem\cite{pearl2016book}:
%\begin{small}
\begin{equation}
    \setlength{\abovedisplayskip}{2pt}
    \setlength{\belowdisplayskip}{1pt}
    %\small
    P(A|do(Q=q),V=v)=\sum_u P(Q=q,V=v,u)P(u)
\end{equation}
%\end{small}
where $do(\cdot)$ is the do-operator (do-calculus)\cite{pearl2016book} and $do(Q=q)$ can be understood as cutting all the original incoming arrows to $Q$, making $Q$ and $U_c$ independent.

%为了进一步减少语言偏差的影响，我们用反事实干预来蒸馏出混淆带来的Q-R之间的伪相关性，从主干模型的因果效应中去除伪相关，从而控制混淆变量Uwx对A的因果效应。
%To further mitigate the influence of the keyword bias, we employ counterfactual generation to distill the spurious correlation introduced by confounding between Q-R, removing it from the causal effect of the main model.
%high-frequency keywords使得模型倾向于学习keywords到答案之间的直接因果效应而不是根据语义和图片信息进行推理，the high-frequency keywords make Q and A spuriously correlated 
In Figure \ref{SCM}b, the syntactic structure and keywords $Q_{s,k}$ influence $A$ by two paths: $Q_{s,k} \rightarrow R \rightarrow A$ and $Q_{s,k} \rightarrow A$. We formulate the spurious correlation between keyword bias and the answer as the direct causal effect of the keyword on the answer, which is the path $Q_{s,k} \rightarrow A$. 
We mitigate the keyword bias by distilling the direct causal effect of the keyword and subtracting it from the total causal effect.
% \begin{equation}
%     \setlength{\abovedisplayskip}{3pt}
%     \setlength{\belowdisplayskip}{3pt}
%     \small
%     equation
% \end{equation}

%当前的研究表明偏差主要存在于语言模态，因此我们主要考虑对语言模态进行因果元素干预。如第3.1节所述，混淆变量的存在使得Q-R的相关性不置信，导致对Q-R之间的因果关系产生了混乱，从而影响答案A。
%因此，CIBi通过应用反事实生成的方式来干预问题当中的语境和语义关键词，然后利用对比学习迫使模型学习问题中真正的语义信息，从而打断了Q-U之间的后门路径。具体的说，对于语境干预，我们通过固定关键词，用相同语义的语境替换问题当中的语境，从而实现对问题中语境的干预。对于关键词干预，我们通过固定语境片段，用同义词对问题当中的关键词进行替换，从而实现对问题中关键词的干预。如图2b所示，阴影部分表示未改变的问题部分。
\subsection{Selection Of Fine-grained Bias}
%keyword、syntactic structure，K+S
In this section, we have provided a detailed explanation of the process for determining the specific concept for each component leading to fine-grained bias, including syntactic structure, keywords, and context.
For example, given a sample with the question "What color is the banana?", we extract question-type words such as "What color" for each question as the syntactic structure, and select "banana" as the keyword. We define the co-occurrence pattern of "what color" and "banana" as the context.
%Fig. 3 shows examples with nine shortcut-specific concepts (see App. B.2 for more examples). 

\textbf{Syntactic Structure (S):} We consider the 65 question prefixes in the annotation of the original VQA dataset as the concept of syntactic structure.

\textbf{Keyword (K): } Following Si\cite{LP} which utilize the mutual information to measure the mutual dependence, for a QA pair $(q,a)$, we calculate the mutual information of each token $e \in q$ in the remaining sentence (except the question-type words) to the ground-truth answer $a$ as:
\begin{equation}
    \label{eq:MI}
    \setlength{\abovedisplayskip}{3pt}
    \setlength{\belowdisplayskip}{3pt}
    \mathcal MI(w,a) = \log \frac{H(w,a)}{H(w)*H(a)/K}
\end{equation}
%Given a sample (v, q, a), (where v ∈ V , q ∈ Q and a ∈ A), we utilize the mutual information to measure the mutual dependence between the answer a and each token w ∈ q in the question sentence (which excludes the question type part). The mutual information of the word5 w and the answer a is:
where $H(w)$, $H(a)$ and $H(w, a)$ respectively represent the total numbers of samples in which $w$, $a$ and their co-occurrence occur. $K$ is the total number of samples in the dataset. Richer mutual information means stronger correlation between the word and the answer. 
we select the word with highest score from the remaining sentence (except the question-type words) as keyword.
%As shown in Fig. 3, we can always find the the most relevant keyword to the answer in the question.

\textbf{Context (C):} We defined the sepcific pattern of question-type words and keyword from the remaining sentence as the concept of context for a given sample.

%Critical Words Selection (CW SEL.) In this step, we first extract question-type words for each question Q2 (e.g., ”what color” in Figure 3). Then, we select top-K words with highest scores from the remaining sentence (except the question-type words) as critical words. The counterfactual question Q− is the sentence by replacing all critical words in Q with a special token “[MASK]”. Meanwhile, the Q+ is the sentence by replacing all other words (except question-type and critical words) with “[MASK]”. We show an example of Q, Q+, and Q− in Figure 3.

\subsection{Context Debiasing}
\label{Sec:context debias}
%Current research indicates that bias primarily exists within the language modality. Therefore, our primary focus lies in conducting causal interventions within the language modality. 
As described in Section \ref{Sec:Structural Causal Model}, the confounding variable undermines the confidence in the correlation between $Q$ and $R$, further affecting the answer $A$. 
Based on the theory of causal intervention, we design a context-debiasing branch to control confounder bias and disrupt the backdoor path $U_{c}\rightarrow Q$ by counterfactual generation and contrastive learning. 
%CIBi intervenes in the context and semantic keywords within the question by employing counterfactual generation. Then, it utilizes contrastive learning to compel the model to learn the genuine semantic information of the question, thus disrupting the backdoor path $U_{c,w} \rightarrow Q$. 
%我们
Specifically, for counterfactual generation, we first maintain a fixed syntactic structure fragment and replace the keywords in the question with synonyms. And then we keep the keywords fixed and replace the syntactic structure in the question with semantically equivalent context. As illustrated in Figure \ref{fig:model}, the shaded portion represents the unchanged part of the question.

%在得到进行了反事实干预的问题表征后，我们在三元组级别（例如，（IQA，IQ+A））上进行对比学习，将VQA中的每个问答（IA）匹配视为一个整体，然后与因果干预问题Q+的IA匹配形成正（IA，Q+）对，而与其未配对问题 Q− 匹配的 IA 形成负对。while a IA matching with its unpaired question Q− forms a negative pair.我们使用常见的归一化温度尺度交叉熵损失(InfoNCE)作为对比学习的目标损失函数，具体如下：

After obtaining the causally intervened question representation, we perform contrastive learning to enhance the interaction between modalities and magnify the differentiation of fusion representation among samples.
%we perform contrastive learning by constructing $(VQ, VQ^+)$ pairs. 
We create positive pairs $(V, Q^+)$ by matching $V$ with its causally intervened question $Q^+$ and negative pairs by matching $V$ with its unpaired question $Q^-$, which are the images within the mini-batch except the matched $Q$. For this purpose, we utilize the commonly used normalized temperature-scaled cross-entropy loss (InfoNCE) as the objective loss function for contrastive learning, formulated as follows:

\begin{equation}
    \label{eq:L_cl}
    \setlength{\abovedisplayskip}{3pt}
    \setlength{\belowdisplayskip}{3pt}
    \mathcal L_{cl} = -\frac{1}{N}\sum_{n=1}^N \mathcal J^{(n)}
\end{equation}
\begin{equation}
    \label{eq:L_cl}
    \setlength{\abovedisplayskip}{3pt}
    \setlength{\belowdisplayskip}{3pt}
    \mathcal J^{(n)} = \frac{1}{K}\sum_{k=1}^Klog\frac{\exp{h(E_{(vq)_n},E_{(vq^+)_{n,k}})/\tau)}}{\sum_{i=1}^N\exp{h(E_{(vq)_n},E_{(vq)_i}))/\tau)}}
\end{equation}
%其中，h表示样本x的表征，h+表示样本x 对应是增强正样本的表征

where $E_{(vq^+)}$ denotes embedding of positive samples during training, $h$ estimates the similarity of two samples, $\tau$ is the temperature parameter, and $K$ is the number of positive samples. In this paper, we use cosine similarity for the estimation.

%之前类似的工作是将对比学习应用在文本表征输出上，我们将Causal-effect Disentanglement module应用在模态融合层的输出上，类似于一个可以即插即用的模块，因此它可以直接套用到现有的视觉问答模型中，作为一个辅助的模块。首先将样本中的文本问题和图片分别输入到文本编码器、图片编码器，得到二者的特征 embedding，然后将文本和图片特征输入到模态融合层，进行模态的融合，得到融合模态的特征 embedding。然后将原样本的融合模态特征和因果干预后的样本的融合模态特征同时输入到Causal-effect Disentanglement module，进行对比损失计算。
%(Making independent)
%Similar prior work applied contrastive learning to text representation outputs. To more precisely disentangle causal effects, we employ contrastive learning at the output of the modality fusion layer. This module acts as a plug-and-play component, allowing it to be seamlessly integrated into existing VQA models as an auxiliary module. 
%In this approach, we start by inputting the textual questions and images separately into the text encoder and image encoder, respectively. This generates feature embeddings for both modalities. Subsequently, these text and image features are fed into the modality fusion layer to combine the modalities, yielding the fused modality feature embedding. Both the original fused modality features of the samples and the features from the causally intervened samples are then input into the Causal-effect Disentanglement module for contrastive loss computation.
Different from previous approaches that applied contrastive learning to textual representation, we use contrastive learning at the modality fusion layer. 
We input both the original fused modality features of the samples and the features from the positive samples and negative samples into the contrastive learning module for computing the contrastive loss. 
%It is worth noting that the Context-debiasing branch can be seamlessly integrated into existing VQA models as a plug-and-play auxiliary module.
%同时也将原样本的融合模态特征经过归一化后，计算交叉熵损失：（equation）总的损失函数是交叉熵损失与对比损失的加权求和，具体公式如下:
% Additionally, the fused modality feature of the original sample is normalized and utilized to compute the cross-entropy loss:
% \begin{equation}
%     \setlength{\abovedisplayskip}{3pt}
%     \setlength{\belowdisplayskip}{3pt}
%     L_{vqa} = -\frac{1}{N}\sum^{N-1}_{i=0}log(softmax(F_{vqa}(v,q))) 
% \end{equation}
% The overall loss function is the weighted sum of the cross-entropy loss and the contrastive loss, expressed by the following formula:
% \begin{equation}
%     \setlength{\abovedisplayskip}{3pt}
%     \setlength{\belowdisplayskip}{3pt}
%     L_{ca} = L_{vqa} + \beta L_{cf}
% \end{equation}

%混淆蒸馏的目的是为了进一步消除语言偏差，使用反事实干预的方法，让模型根据有偏的观察做出无偏的预测。具体来说，我们首先使用有偏训练集训练一个有偏差的模型，然后通过掩码部分关键词，作为反事实样本输入到训练好的模型提取关键词偏差。使用Q表示原始问题，Qm表示掩码关键词后的问题，fqa表示有偏模型的输出，则去除关键词偏差的因果效应可以表示为：

\subsection{Syntactic Structure and Keyword Debiasing}
\label{Sec:context debias}
In this section, we specifically introduce how to reduce keyword bias from the newly designed question-only branch called SK-debiasing branch. Technically, the question-only branch is added alongside the base VQA model. We generate counterfactual samples by randomly masking keywords and syntactic structure, and then input them into the question encoder $E_q$ to extract question embedding $q^*_i$. Then we distill and calculate keyword bias by a mapping $\mathcal F_{qa}(q^*_i)$, which is a neural network consisting of three linear layers and a sigmoid function. 
%我们用一个融合函数来
%The output scores are fused by the function h to obtain the final score Zq,v,k
%The $F_{qa}(q^*_i)$ and $F_{vqa}(v_i,q_i)$ are fused by the function h to obtain the output $F_^*_{vqa}(v_i,q_i)$, which is formalized as: 
We utilize fusion function $\mathcal H$ to fuse $\mathcal F_{qa}(q^*_i)$ and $\mathcal F_{vqa}(v_i,q_i)$:
%因为我们依靠f进行推理，所以the distribution of
%Since we mainly rely on $\mathcal{F}_{vqa}$ for prediction, the distribution of $\mathcal{F}_{vqa}$, $\mathcal{F}_{qa}$ can well represent the causal effect of $\langle V,Q \rangle$ and keyword bias.

\begin{equation}
    \setlength{\abovedisplayskip}{3pt}
    \setlength{\belowdisplayskip}{3pt}
    %\small
    \mathcal{F}^*_{vqa}(v_i,q_i) = \mathcal{H} (\mathcal{F}_{vqa}(v_i,q_i),\mathcal{F}_{qa}(q^*_i)) 
\end{equation}
\begin{equation}
    \setlength{\abovedisplayskip}{3pt}
    \setlength{\belowdisplayskip}{3pt}
    \mathcal{H} (\mathcal{F}_{vqa}(v_i,q_i),\mathcal{F}_{qa}(q^*_i)) = \log \sigma (\mathcal{F}_{vqa}(v_i,q_i) - \mathcal{F}_{qa}(q^*_i))
\end{equation}
%To measure the causal effect of keyword bias in the VQA models, one effective approach is to train a simple model using the masked question modality alone. 
%我们输入到训练好的模型提取关键词偏差

%In order to avoid the instability brought by the adversarial scheme, the question-only model is trained with the objective of minimizing the standard cross entropy loss, which is formalized as:
The training strategy follows RUBi\cite{RUBi}. The base VQA model and question-only branch are trained with the objective of minimizing the standard cross entropy losses, which are formalized as follows:
%over the scores Zq,v,k, Zq and Zv:

\begin{equation}
    \label{eq:L_qa}
    \setlength{\abovedisplayskip}{3pt}
    \setlength{\belowdisplayskip}{3pt}
    \mathcal{L}_{QA} = -\frac{1}{N}\sum^{N-1}_{i=0}log(softmax(\mathcal{F}_{qa}(q^*_i))) [A_i]
\end{equation}
\begin{equation}
    \label{eq:L_vqa}
    \setlength{\abovedisplayskip}{3pt}
    \setlength{\belowdisplayskip}{3pt}
    \mathcal{L}^*_{VQA} = -\frac{1}{N}\sum^{N-1}_{i=0}log(softmax(\mathcal{F}^*_{vqa}(v_i,q_i))) [A_i]
\end{equation}
%The purpose of confounder distillation is to further alleviate language bias using counterfactual generation, enabling the model to make unbiased predictions based on biased observations. Specifically, we first train a biased model on a biased training dataset. Then, by masking certain keywords, we use these counterfactual samples as inputs to the trained biased model to extract keyword bias. Using $E$ to denote the causal effect, $q$ to denote the original question, $q^*$ to denote the question with masked keywords, $F_{vqa}(V=v,Q=q)$ to represent the output of the VQA model, and $F_{q^*a}(Q=q*)$ to represent the output of the biased model, the causal effect for mitigating keyword bias can be represented as follows:

%很明显根据反事实分支得到的概率越大，说明这个答案包含的语言偏差越大。
%和RUBi一样这个分支只在训练的时候用到，测试的时候把它去掉。
%Similar to RUBi, this branch is only used during training and is removed during testing.
\subsection{Model Training and Testing Strategy}

\label{Sec:Training}
%We jointly optimize the parameters of the base VQA model, context-debias branch and keyword-debias branch in an end-to-end training scheme. The main loss $\mathcal L^*_{VQA}$ refers to the cross-entropy loss associated with the predictions of $\mathcal F^*_{vqa}(v_i,q_i)$ from Equation \ref{eq:L_vqa}.
%The loss $\mathcal {L}_{CL}$ denotes the contrastive learning from Equation \ref{eq:L_cl}. The question-only loss $\mathcal {L}_{QA}$ is a cross-entropy loss associated with the predictions of $\mathcal F_{qa}(q_i)$ from Equation \ref{eq:L_cl}.
In general, we construct our total loss $\mathcal L_{total}$ by summing the $\mathcal L_{CL}$, $\mathcal L_{VQA}^*$ and $\mathcal L_{QA}$ together in the following equation:

\begin{equation}
    \setlength{\abovedisplayskip}{3pt}
    \setlength{\belowdisplayskip}{3pt}
    \mathcal L_{total} = \lambda \mathcal L_{CL} + \mathcal L^*_{VQA} + \mathcal L_{QA}
\end{equation}
where $\mathcal L^*_{VQA}$, $\mathcal L_{QA}$ are over $\mathcal F^*_{vqa}(v_i,q_i)$, $\mathcal F_{qa}(q_i^*)$, $\mathcal L_{CL}$ represents the objective loss function for contrastive learning, $\lambda$ is a weighting hyperparameter. We backpropagate $\mathcal L_{total}$ to optimize all the related parameters, including the parameters of the base VQA model, the context-debiasing branch, and the question-only branch. Additionally, we prevent the gradients computed from $\mathcal L_{QA}$ from being propagated back to the question encoder to avoid direct capture of keyword bias. 
%branch只用在训练阶段，目的是从有偏的数据中训练得到无偏的模型。我们base VQA model 在VQA-cp数据集上测试。
%After training, the question-only branch is removed to maintain the original VQA model unchanged.
%The two branches are employed only during the training phase to learn an unbiased model from biased data. When testing, we remove both of them to maintain the base VQA model unchanged.
The two branches are employed only during training to learn an unbiased model from biased data. When testing, we remove both of them to maintain the base VQA model unchanged.

\section{EXPERIMENT}
\label{EXPERIMENT}
\subsection{Datasets and experiment settings}
\label{Datasets}
%We use VQA-CP v2[cite] dataset to evaluate the effectiveness of our method. VQA-CP v2 dataset was built by reorganizing VQA v2, which is designed to test the robustness of the VQA models. The answer distribution of its training set and test set is quite different, so it can usually be used to evaluate the robustness of the VQA models. We follow the protocol of3,21 to train and evaluate our model, and use the standard VQA evaluation metrics to evaluate the performance of our model.
%对比损失在整体损失中的权重调整参数beta，本文将其设置为了0.4，每一个批次的训练样本数量是 256，训练轮次是27。其他的实验设置，全都和基线模型保持一致。
%question-mask branch
We evaluate the effectiveness of our method on VQA-CP v2\cite{VQACP}. VQA-CP v2 was built by reorganizing VQA v2\cite{VQAv1}, which is designed to test the robustness of the VQA models. Additionally, we present results on the balanced VQA v2 dataset to assess whether our method over-corrects language bias. 

%We conduct experiments with three baseline VQA architectures: Stacked Attention Network (SAN) [50], Bottom-up and Top-down Attention (UpDn) [5], and a simplified MUREL [10] (S-MRL) [11].
The models are evaluated via standard VQA evaluation metric\cite{VQAv1}: the accuracy of all answers and different question types. 
We conduct experiments based on the following baseline models: RUBi\cite{RUBi}, LPF\cite{LPF}, CF-VQA\cite{CF-VQA}, CSS\cite{CSS}. 
The weight adjustment parameter $\lambda$ in the loss $\mathcal L_{total}$ is set to 0.4 with sensitivity analysis provided(Figure \ref{fig:dt}). 
%The weight adjustment parameter $\lambda$ in the loss $\mathcal L_{total}$ is set to 0.4 in this paper. 
Each training batch consists of 256 samples, and there are 30 training epochs. All other experimental settings remain consistent with the baseline model.

\begin{table*}[!t]
\begin{center}
\caption{Comparison of the accuracies (\%) on VQA-CP v2 $test$ and VQA v2 $val$ set with different baselines. }\label{table1}
\label{table1}
\setlength{\abovecaptionskip}{0.1cm}  
\setlength{\belowcaptionskip}{-0.1cm}  
\centering
\tiny
\renewcommand{\arraystretch}{1.0}
\resizebox{0.98\linewidth}{!}{
\begin{tabular}{ccccccccccc}
\Xhline{1px}
\multirow{3}{*}{Models} & \multirow{3}{*}{Base} & \multicolumn{4}{c}{VQA-CP v2 test} & \multicolumn{4}{c}{VQA v2 val} & \multirow{3}{*}{Gap$\Delta$}\\ 
\cmidrule(lr){3-6} \cmidrule(lr){7-10}

& & All & Y/N & Num & Other & All & Y/N & Num & Other \\
\hline
SAN\cite{SAN} & - & 24.96 & 38.35 & 11.14 & 21.74 & 52.41 & 70.06 & 39.28 & 47.84 & - \\
UpDn\cite{UpDn} & - & 39.74 & 42.27 & 11.93 & 46.05 & 52.41 & 70.06 & 39.28 & 47.84 & - \\
S-MRL\cite{RUBi} & - & 38.46 & 42.85 & 12.81 & 43.20 & 63.10 & - & - & - & - \\
%\hline
%\multicolumn{10}{l}{\emph{methods based on weakening language prior follow:}} \\
\hline
RUBi\cite{RUBi} & S-MRL & 47.11 & 68.65 & 20.28 & 43.18 & 61.16 & - & - & -& - \\
%LM & - & 48.78 & 72.78 & 14.61 & 45.58 & 63.26 & 81.16 & 42.22 & 52.22 \\
LPF\cite{LPF} & UpDn & 55.34 & 88.61 & 23.78 & 46.54 & 55.01 & 64.87 & 37.45 & 52.08 & - \\
CF-VQA\cite{CF-VQA} & S-MRL & 55.05 & 90.61 & 21.50 & 45.61 & 60.94 & 81.13 & 43.86 & 50.11 & - \\
CSS\cite{CSS} & UpDn & 58.95 & 84.37 & 49.42 & 48.21 & 59.91 & 73.25 & 39.77 & 55.11 & - \\
%CIBi (ours) & LM & - & - & - & - & - & - & - & - \\
%\hline
%\multicolumn{10}{l}{\emph{methods based on augmentation follow:}} \\
\hline
CIBi (ours) & RUBi &{49.62}&{72.56}&{31.61}&{44.52}& 61.22 & 75.49 & 40.07 & 53.86 & + 2.51 \\
CIBi (ours) & LPF & 56.71 & 89.23 & 30.98 & 46.59 & 58.41 & 71.03 & 39.18 & 52.61 & + 1.37 \\
CIBi (ours) & CF-VQA & 57.62 & 90.54 & 41.63 & 44.31 & \textbf{61.59} & 82.02 & 43.60 & 51.79 & + 2.19 \\
CIBi (ours) & CSS & \textbf{59.58} & 86.94 & 49.98 & 50.24 & 60.47 & 81.04 & 42.94 & 50.02 & + 0.63 \\
\hline
\multicolumn{10}{l}{Gap$\Delta$ represents the accuracy improvements of the “All" question type on VQA-CP v2 dataset.}
\end{tabular}}
\end{center}
\end{table*}

\begin{table}[!t]
\caption{Ablation study of the CIBi on VQA-CP v2 with baseline RUBi.}\label{tabel2} 
\setlength{\abovecaptionskip}{0.2cm}  
\setlength{\belowcaptionskip}{-0.1cm} 
\centering
\scriptsize
\renewcommand{\arraystretch}{1.2}
\resizebox{1\linewidth}{!}{
\begin{tabular}{ccccc}
\Xhline{1px}
{Models} &{All}&{Y/N}&{Num}&{Other}\\
\hline
{RUBi\cite{RUBi}}&{47.11}&{68.65}&{20.25}&{43.18}\\
{w/o SK-debiasing brach}&{48.37}&{71.34}&{22.43}&{43.06}\\
{w/o context-debiasing brach}&{48.59}&{70.62}&{27.57}&{43.51}\\
\hline
{CIBi}&{49.62}&{72.56}&{31.61}&{44.52}\\
\hline
\end{tabular}}
\end{table}

\subsection{Results and analysis}
\label{results and analysis}
\textbf{Quantitative analysis.} 
%In Table 1, we compare the proposed method with the strong baselines on the VQA-CP v2 dataset. 
% To be fair, We compare the proposed method with several representative VQA approaches which mainly focus on textual modality. RUBi\cite{RUBi}, LPF\cite{LPF}, and CF-VQA\cite{CF-VQA} mitigate language bias by a separate question-only branch. CSS\cite{CSS} balance training data proposed to change the training distribution for unbiased training. 
%To be fair, we only compare our method to approaches that focus on textual modality, which can be grouped as follows: (1)Methods that weaken language bias by a separate question-only branch, including RUBi\cite{RUBi}, LPF\cite{LPF}, and CF-VQA\cite{CF-VQA}. (2)Methods that balance training data proposed to change the training distribution for unbiased training, including CSS. 
%In Table \ref{table1}, we compare CIBi with the strong baselines on VQA-CP v2 and VQA v2. 
For the sake of fairness, we select several representative VQA models as the strong baselines for comparison, all of which mainly focus on the textual modality. RUBi\cite{RUBi}, LPF\cite{LPF}, and CF-VQA\cite{CF-VQA} mitigate language bias by a separate question-only branch. CSS\cite{CSS} balances training data so as to change the training distribution for unbiased learning. 
%The results on VQA-CP v2 and VQA v2 are reported in Table 1. 
Based on the baseline RUBi\cite{RUBi}, our approach reaches an average overall accuracy of 49.62\% with an obvious improvement (+2.51\%). In addition, when built upon LPF\cite{LPF} and CF-VQA\cite{CF-VQA}, CIBi achieves an accuracy gain of +1.37\% and +2.19\%, respectively. By utilizing CSS\cite{CSS} as the baseline, we achieve a relatively slight improvement (+0.63\%). Nevertheless, the results still indicate that our method contributes to discerning fine-grained biases, thereby enhancing the accuracy of “Yes/No” questions and “other” questions. Note that CIBi surpasses the baseline models RUBi\cite{RUBi}, LPF\cite{LPF}, CSS\cite{CSS} by an obvious margin (+3.91\%, +0.71\% and +2.57\%) on “Yes/No” questions and surpasses the baseline models RUBi\cite{RUBi}, LPF\cite{LPF}, CF-VQA\cite{CF-VQA} (+11.33\%, +7.20\% and +20.13\%) on “Number” questions. Because these two types of question are susceptible to fine-grained context bias and keyword bias, which is ignored by previous methods. 
%我们的实验数据和baseline呈现较强的相关性，这是因为我们的方法对模型自身对语言偏差的捕获有一定的依赖。
Additionally, the results show a certain correlation with the baseline. This is because our method to some extent relies on the model's inherent ability to capture language bias.
%Based on the baseline RUBi, Our CIBi approach reaches an average overall accuracy of 49.62\% with a obvious improvement(+2.51) based on RUBi. In addition, our ECBi approach, which builds upon the baseline models LPF and CF-VQA, achieves an accuracy gain of +0.95 percentage points over LPF and +1.19 over CF-VQA. 
%在CSS作为baseline上提升比较少，是因为CSS用了数据增强的方法，但是我们的方法还是有提升的，说明我们的方法去除了他们之前没有注意到的细粒度偏差,contribute to other类的提升。
%Based on the baseline CSS, we reaches a relatively slight improvement(+0.53) because CSS use data augmentation techniques. But the result still indicates our method notice the fine-grained bias, contribute to the improvement in "Yes/No" questons and "other" questions. 
%Note that ECBi surpasses RUBi, LPF, CSS by a obvious margin(+3.91, +0.71, +2.62) on "Yes/No" questons and surpass RUBi, LPF, CF-VQA by a large margin(+5.33, +8.20, +33.20) on "Number" questions, because these two type of questions are susceptible to the influence of fine-grained context bias and keyword bias, which is ignored by previous methods. 
%我们在VQAcp数据集上数据表现良好，说明我们没有过度去偏。
The competitive performance on VQA-CP v2 shows that our approach mitigates the fine-grained language bias on answers effectively and integrates well with other VQA models. Additionally, the results on VQA v1 also demonstrate the competitive performance and robustness of our method, indicating that we have not overly mitigated bias.

\begin{figure}[!t]
    \centering
    \includegraphics[width=0.98\columnwidth]{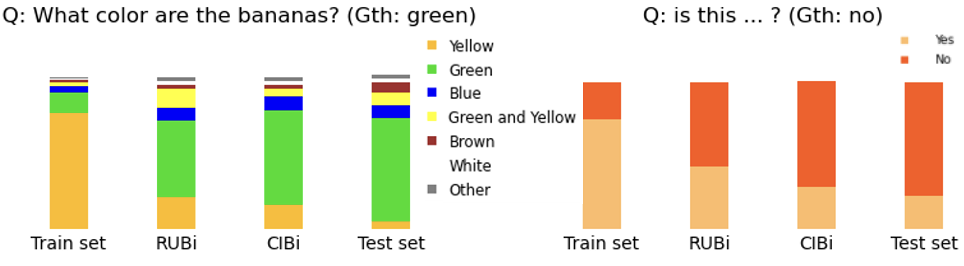}
    \caption{The answer distributions on the VQA-CP v2. CIBi is based on RUBi.}
    \label{fig:dt}
\end{figure}

\textbf{Qualitative analysis.}
%Fig. 3 shows answer distributions for VQA-CP v2 train and test sets, the UpDn and the LPF of two question examples: “What color are the bananas?” and “Does this have lettuce?”. It shows that the original UpDn model tends to overfit the answer distribution of training set and predicts the frequently-used answer in it rather than reasoning from the visual content. In contrast, the LPF can well recover the answer distribution in the test set, which demonstrates that LPF enables the model to be more grounded on the image content.
%The qualitative results are provided to validate whether CF-VQA can effectively reduce language bias and retain language context. As illustrated in Figure 7, CF-VQA can successfully overcome language bias on yes/no questions compared to RUBi, while the baseline model suffers from the memorized language prior on the training set. Besides, for “what kind” questions, RUBi prefers the meaningless answer “none” rather than specific ones.
%The answer distributions for the VQA-CP v2 training and testing sets, along with the RUBi and the LPF of two question examples, are provided to validate the effectiveness of CIBi in mitigating finer-grained language bias.
Figure \ref{fig:dt} illustrates answer distributions for VQA-CP v2 train and test sets, the RUBi and the CIBi of two question examples: “What color are the bananas?” and “is this...?”. For “is this” questions, RUBi tends to overfit the answer distribution of the training set and predicts the frequently-used answer “yes”. Compared to RUBi, our method can better recover the answer distribution in the test set on both question examples, which validates the effectiveness of CIBi in mitigating finer-grained language bias. 
%The qualitative results are provided to validate whether CIBi can effectively reduce finer-grained language bias. As illustrated in Figure \ref{fig:dt}, CF-VQA can successfully overcome language bias on yes/no questions compared to RUBi, while the baseline model suffers from the memorized language prior on the training set.
%Figure \ref{Sec:dt} shows answer distributions for VQA-CP v2 train and test sets, the RUBi and the CIBi based on the RUBi of two question examples: “What color are the bananas?” and “Does this have lettuce?”. The answer distributions show that 
%
%In contrast, the LPF can well recover the answer distribution in the test set, which demonstrates that LPF enables the model to be more grounded on the image content.

%\label{fig:param}
\begin{figure}[!t]
    \centering
    \includegraphics[width=0.95\columnwidth]{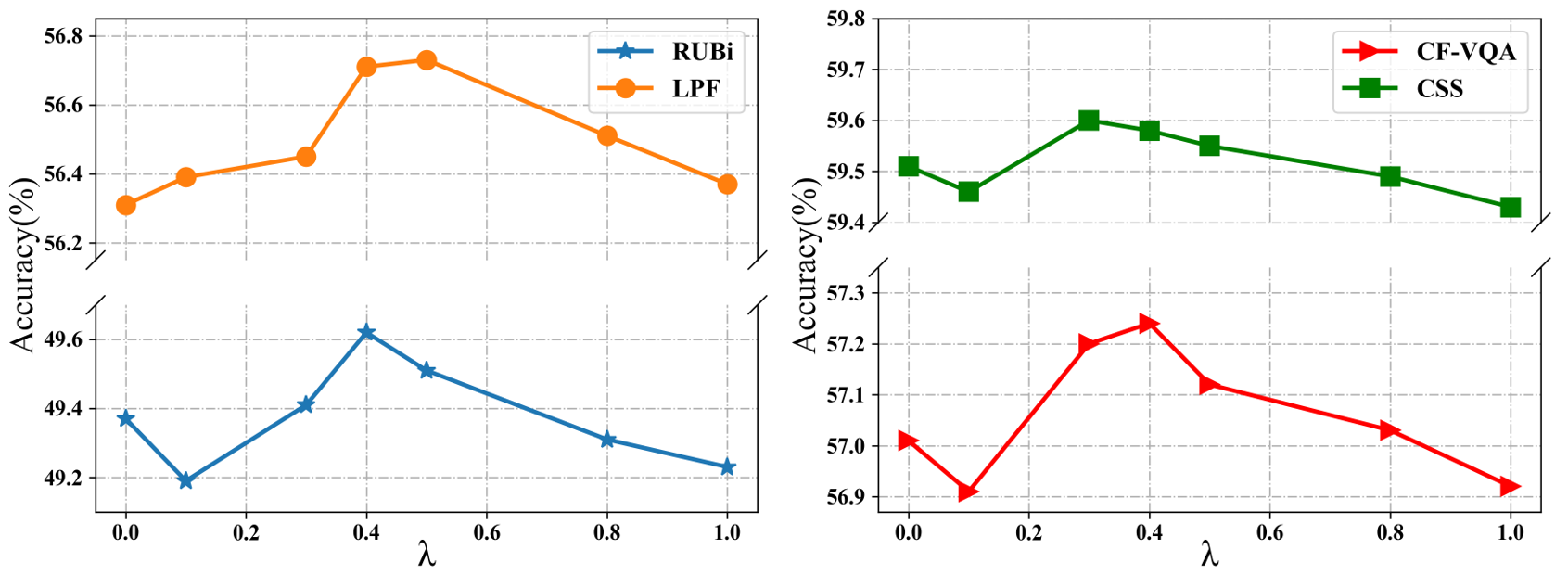}
    \caption{Sensitivity of VQA accuracy on VQA-CP v2 test.}
    \label{fig:param}
\end{figure}

%我们使用因果干预，消除了Q-R之间的伪相关性。
\subsection{Ablation Studies}
\label{Sec:ablation}
%In table \ref{tabel2}, we further conduct ablation studies to validate the effectiveness of the causal-effect adjustment. CF-VQA outperforms 0.93\% for RUBi and 1.07\% for LPF in total accuracy. As we discussed in Section \ref{Sec:Structural Causal Model}, the reason is that we employ causal intervention to eliminate the spurious correlation between $Q$ and $R$.
To validate the effectiveness of the key components in CIBi, we re-train different versions of our model by ablating certain components. The results on VQA-CP v2 are listed in Table \ref{tabel2}. 
Compared to Baseline, $\textit{w/o context-debiasing}$ branch and $\textit{w/o keyword-debiasing}$ branch respectively achieved an overall accuracy increase of +1.26\% and +1.48\%, but still performs worse than the complete model.

%表格分区：backbone；language prior；methods based on augmenttation

\section{CONCLUSION}
\label{CONCLUSION}
%We proposed a finer-grained cross-modal contrastive learning method that jointly trains with a supervised learning objective for robust reasoning. Experiments in different settings are conducted to show that our method promotes reasoning ability in cases where shortcuts are less likely to handle. The proposed learning strategy encourages a soft alignment between vision and language modalities and extracts more abstract knowledge by distinguishing contrastive samples. In this paper, we present that joint training with a vanilla contrastive learning strategy contributes to robust reasoning. However, a main limitation of this paper is that the latest techniques in contrastive learning are not involved and should be further explored. In future works, we hope that self-supervised methods can not only eliminate the shortcut issue but also improve the performance on all categories of questions.
In this work, we propose CIBi, a casual intervention training scheme designed to be model agnostic, to reduce fine-grained language bias learned by the VQA model. It is based on two debiasing branches that capture and remove syntactic structure, keyword and context bias from the textual modality, respectively. Experiments show that CIBi is effective with different kinds of common VQA models. 
%In future work, we will explore the applicability of our approach in addressing unimodal biases across various multi-modal tasks.
%In future works, we would like to extend our approach on other multimodal tasks.
In future work, we will explore the applicability of our learning scheme in addressing uni-modal biases across various multi-modal tasks.

\bibliographystyle{IEEEbib}
\bibliography{refs}

\end{document}